\documentclass[runningheads]{llncs}

\usepackage{eccv}

\usepackage{eccvabbrv}

\usepackage{graphicx}
\usepackage{booktabs}

\usepackage{marvosym}
\usepackage{multirow}
\usepackage{colortbl}
\usepackage{wrapfig}

\usepackage[accsupp]{axessibility}  

\usepackage{hyperref}

\usepackage{orcidlink}

\begin{document}

\title{RT-Pose: A 4D Radar Tensor-based 3D Human Pose Estimation and Localization Benchmark} 

\titlerunning{RT-Pose}

\author{Yuan-Hao Ho\inst{1*}\orcidlink{0000-0003-1089-5509} \and
Jen-Hao Cheng\inst{2*}\orcidlink{0000-0002-6970-3738} \and
Sheng Yao Kuan\inst{2}\orcidlink{0009-0001-5054-3033} \and
Zhongyu Jiang\inst{2}\orcidlink{0000-0003-4462-6497} \and
Wenhao Chai\inst{2}\orcidlink{0000-0003-2611-0008}
Hsiang-Wei Huang\inst{2}\orcidlink{0009-0009-2474-8869} \and
Chih-Lung Lin\inst{1} \and \\ 
Jenq-Neng Hwang\inst{2}\orcidlink{0000-0002-8877-2421}
\\ * Indicates equal contribution
}

\authorrunning{Y.~Ho et al.}

\institute{ National Cheng Kung University \and
University of Washington
\\
\email{n28081527@gs.ncku.edu.tw\textsuperscript{1},\\
\{andyhci, shengyao, zyjiang, wchai, hwhuang, hwang\}@uw.edu\textsuperscript{2}, \\
cllin@ee.ncku.edu.tw\textsuperscript{1}
}
}

\maketitle
\begin{abstract}
Traditional methods for human localization and pose estimation (HPE), which mainly rely on RGB images as an input modality, confront substantial limitations in real-world applications due to privacy concerns. In contrast, radar-based HPE methods emerge as a promising alternative, characterized by distinctive attributes such as through-wall recognition and privacy-preserving, rendering the method more conducive to practical deployments. This paper presents a Radar Tensor-based human pose (RT-Pose) dataset and an open-source benchmarking framework. RT-Pose dataset comprises 4D radar tensors, LiDAR point clouds, and RGB images, and is collected for a total of 72k frames across 240 sequences with six different complexity level actions. The 4D radar tensor provides raw spatio-temporal information, differentiating it from other radar point cloud-based datasets. 
We develop an annotation process, which uses RGB images and LiDAR point clouds to accurately label 3D human skeletons.
In addition, we propose HRRadarPose, the first single-stage architecture that extracts the high-resolution representation of 4D radar tensors in 3D space to aid human keypoint estimation.
HRRadarPose outperforms previous radar-based HPE work on the RT-Pose benchmark. 
The overall HRRadarPose performance on the RT-Pose dataset, as reflected in a mean per joint position error (MPJPE) of $9.91 cm$,  indicates the persistent challenges in achieving accurate HPE in complex real-world scenarios. RT-Pose is available at \url{https://huggingface.co/datasets/uwipl/RT-Pose}.

\end{abstract}

\section{Introduction}
Human localization and 3D pose estimation~\cite{chai2023global,zhang2023mpm,jiang2023unihpe,liu2023posynda,jiang2024back,zhou2024efficient,jiang20242d,yang2023camerapose} are indispensable in Augmented/Virtual Reality~\cite{guzov2021human, lin2010augmented, 9417791, chai2023stablevideo}, human-computer interaction~\cite{yuan2021simpoe, hgr}, healthcare~\cite{NEURIPS2022_af9c9c6d, chen2018patient, lin2021innovative}. 
However, capturing human poses across diverse scenarios and generating comprehensive 3D representations for individuals in varying poses present significant challenges. Camera based motion capture systems are often used to collect human body posture~\cite{ionescu2013human3,wang2021deep,lin2020innovative}. However, these systems are sensitive to light and require careful multi-camera calibration, making them unreliable for outdoor scenarios and unable to provide overall pose estimation~\cite{chen2022mmbody}. 
Furthermore, due to privacy concerns, deploying its application in home or long-term care center scenarios poses challenges.

Radar-based method emerges as a promising alternative, characterized by distinctive attributes such as through-wall recognition~\cite{zhao2018through, zheng2021human}.
In addition, radar is robust to lighting conditions and resilient to various weather conditions and occlusion~\cite{adib2015rf, xie2023rpm}. 
These characteristics make it especially suitable for safety- and privacy-critical applications~\cite{ahuja2021vid2doppler,sun2022human}.
In smart automotive applications, radar, as a complementary sensing modality, can complement adverse scenarios such as low-lighting environments and adverse weather~\cite{cross_check}, challenging RGB-based sensing.
In healthcare applications, RGB-based Human Pose Estimation (HPE)poses privacy risks and is hindered by occlusion, prompting a demand for radar in indoor care environments. 
Considering the advantages and prospects of radar technology, it is indispensable to provide a dataset that contains divwerse scenarios for fostering radar-based HPE research.

The previous radar-based HPE methods~\cite{xue2021mmmesh,sengupta2020mm,endo2023multi} often utilize Constant False Alarm Rate (CFAR) techniques \cite{de2007design,cruw3d,wang2021rodnet} to extract radar point clouds. However, the effectiveness of CFAR-based point cloud extraction can be easily influenced by variations in radar module types and hardware parameters. Furthermore, different human body reflections in radar signals across diverse environments may require specific adjustments to CFAR parameters, making it challenging to generalize their applications. 
To address this issue, some works decompose the 4D radar tensor into vertical and horizontal directions as the input of the model to estimate human pose \cite{lee2023hupr,xie2023rpm,yu2023mobirfpose}. This approach computes the magnitude of radar signals for both directions, which effectively reduces data loss during the conversion to point cloud. However, a 4D radar tensor with Doppler values contains the velocity information, which is more informative than a decomposed radar tensor~\cite{paek2022k, cheng2023centerradarnet}. Therefore, this work focuses on exploring the use of raw 4D radar tensors for HPE and establishing a benchmark based on this data format. 

The presented radar tensor-based human pose (RT-Pose) dataset is the first benchmark to integrate calibrated 4D radar tensors, RGB images, and LiDAR point clouds data in the field of HPE. Releasing a multi-modality dataset marks a significant advancement, particularly in the domain of human behavior analyses. There are 
a total of 72k frames in 240 sequences in the dataset. The actions are organized into sequences of increasing complexity, providing a realistic motion distribution. All experiments are conducted in 5 environmental conditions in 8 scenarios, which ensures the variety of scenarios for better comprehension of the model. Along with the release of this comprehensive RT-Pose dataset, we also build upon the High-Resolution Network (HRNet)~\cite{wang2020deep} to propose  HRRadarPose model, a robust baseline for 3D HPE utilizing 4D radar tensor data. 
Furthermore, our study also includes a comparative evaluation of different radar data types within our proposed model, as well as benchmarking against previous radar-based methods. These results demonstrate the advantages of employing the 4D radar tensor method over traditional radar point cloud methods, especially in complex actions and diverse scenarios. The 4D radar tensor-based approach presents a challenging but promising direction for future research. In summary, our contributions include:
\begin{itemize}
    \item The first dataset provides calibrated 4D radar tensors, LiDAR point clouds, and RGB images for human localization and 3D pose estimation of complex actions in diverse scenarios.
    \item We develop a reliable 3D human pose annotation workflow by joint optimization with LiDAR point cloud and RGB images for both outdoor and indoor scenarios.
    \item We propose HRRadarPose, the first single-stage human pose estimator specifically designed to learn 4D radar tensors' representation for the human pose estimation task.

\end{itemize}

\section{Related Works}
\begin{table*}[t]
\centering
\caption{Comparison of 3D human pose estimation datasets. The first group of rows shows RGB and LiDAR datasets. The second group of rows presents datasets with radar. The radar point cloud is denoted as RPC, and the 4D radar tensor is denoted as 4D-RT.
$*$ indicates a modified 4D radar tensor with a specific velocity range.}
\resizebox{\textwidth}{!}{
\begin{tabular}{l|cccccccccc}
\toprule
Dataset & RGB           &Depth& LiDAR & RPC& 4D-RT& \# Scenes            & \# Actions                 & \# Seqs & \# FPS                    & \ Time (min) \\ \midrule
Human3.6M \cite{ionescu2013human3} & $\checkmark$  &&  &                                        &                                       &  - & 15  & 839          &     50                    & 1200   \\
HSC4D \cite{dai2022hsc4d}                       & $\checkmark$  && $\checkmark$         &                                        &                                       & 3                    & 20 &     -      & 20 & 42      \\
SLOPER4D \cite{dai2023sloper4d}                  & $\checkmark$  && $\checkmark$         &                                        &                                       & 10                   & -                        & 15          &      20                    & 83      \\
\midrule
MARS \cite{an2021mars}                       & &&  & $\checkmark$       &                                       & 1                    & 10& 80& 10& 80\\
mRI \cite{NEURIPS2022_af9c9c6d}                       & $\checkmark$  &$\checkmark$       &  & $\checkmark$       &                                       & 1                    & 12                         & 300          & 10                        & 264     \\
HuPR \cite{lee2023hupr}                        & $\checkmark$  &&  &                                        & $\checkmark*$     & -                  & 3                          & 235           &         10                & 240      \\
\midrule
RT-Pose~(Ours) & $\checkmark$  && $\checkmark$         &       & $\checkmark$      &     40                 & 6                         & 240          &     10                    &  120       \\ \bottomrule
\end{tabular}
}
\label{tab:dataset}
\end{table*}
\subsection{3D Human Pose Estimation Datasets}
As one of the fundamental tasks in the computer vision field, there are lots of works introducing datasets and benchmarks for 3D HPE, as shown in Table~\ref{tab:dataset}. Human3.6M~\cite{ionescu2013human3} is the first large-scale dataset for 3D human sensing in lab environments, which contains 3.6-million frames of corresponding 2D and 3D human poses from mocap captured videos of 5 female and 6 male subjects. Compared to Human3.6M, MPI-INF-3DHP~\cite{mehta2017monocular} is a more challenging 3D human pose dataset captured in the wild. There are 8 subjects with 8 actions captured by 14 cameras covering a greater diversity of poses. 3DPW~\cite{von2018recovering},  the first one to include video footage taken from a moving phone camera, includes 60 video sequences. Recently, several works have explored the possibility of HPE from other modalities like radar and LiDAR. mmBody~\cite{chen2022mmbody} consists of synchronized and calibrated mmWave radar point clouds and RGB-D images in different scenes, as well as skeleton/mesh annotations for humans in the scenes. 

mRI~\cite{NEURIPS2022_af9c9c6d}, a multi-modal 3D HPE dataset using mmWave, RGB-D, and inertial sensors, generates a radar point cloud dataset based on the CFAR method. There are 12 rehabilitation exercises selected to evaluate radar application in home-based health monitoring. 
HuPR~\cite{lee2023hupr} collects 235 sequences of data by calibrated radar and camera for 2D HPE in an indoor environment. 

All open datasets for radar-based human pose estimation are collected within a limited detection area, which contradicts the characteristics of radar applications for tracking. Although detecting within a small area may ensure system accuracy, it limits the system's practical application.
Furthermore, to the best of our knowledge, our dataset is the first to include radar, LiDAR, and RGB sequence information with 3D pose annotations. This allows our dataset to provide pose annotations for both indoor and outdoor environments, distinguishing it from existing datasets.

\subsection{Radar-based Human Pose Estimation}
In spite of the popularity of RGB-based HPE applications, privacy concerns have driven the exploration of radio or radar modalities for HPE. RF-Pose~\cite{zhao2018through} is a pioneering application in 2D HPE utilizing radio frequency signals. RF-Pose demonstrates the feasibility of pose detection in through-wall scenarios and provide the possibility of predicting the 3D pose using radio signals.

On the other hand, frequency-modulated continuous wave (FMCW) radar, which operates in the millimeter-wave band (30-300GHz), has been shown to be suitable for detection and HPE tasks. 
The radar point cloud is widely used to represent the radar signal in HPE~\cite{an2021mars,sengupta2020nlp,zhang2021comprehensive,NEURIPS2022_af9c9c6d,xue2021mmmesh,chen2022mmbody,sengupta2020mm}. mmMesh~\cite{xue2021mmmesh} introduces a temporal model structure that concurrently captures global and local 3D point cloud structures in spatial dimensions, ensuring precise localization performance for pose estimation. Nonetheless, in cluttered environments, the probability and intensity of radar signal reflections from the human body decrease. Consequently, implementing point cloud-based methods in real-world applications becomes challenging.

Compared to the radar point cloud methods, 
utilizing 4D tensor radar signal proves to be more informative and reliable
~\cite{xie2023rpm,zhao2018rf,lee2023hupr,zhao2022angle}. RF-pose 3D~\cite{zhao2018rf} is the first work using the 4D RF tensor to predict 3D skeletons in 22 different locations. However, the used RF system is not commercially available, which is hard to reproduce. HuPR~\cite{lee2023hupr} utilizes two pieces of well-calibrated TI mmWave radar modules to capture vertical and horizontal radar signals, enhancing the resolution of radar elevation signals to improve pose estimation performance. RPM2.0~\cite{xie2023rpm} adopts a similar hardware setup to HuPR but utilizes the HRNet~\cite{wang2020deep} as the backbone and includes an attention model to reconstruct missing keypoints. However, their two-stage model is trained by only walking pose, restricting the actions' complexity for achieving better pose estimation performance. 
The methods proposed in HuPR and RPM2.0 fuse information from both radar modules.
While the dual-radar-module configuration potentially captures fine-grained information, its intricate setup poses reproducibility challenges. 

 Unlike previous works, our system employs a single radar module for vertical and horizontal sensing, simplifying the sensors' calibration and synchronization. We use 4D radar tensor as training data for 3D HPE, reducing the likelihood of data loss during radar point generation. Moreover, RT-Pose dataset includes complex actions and is recorded in diverse scenes, expanding the applicability to real-world scenarios.

\section{RT-Pose Dataset} \begin{wrapfigure}{b}{0.5\linewidth}
\centering
\includegraphics[width=1\linewidth]{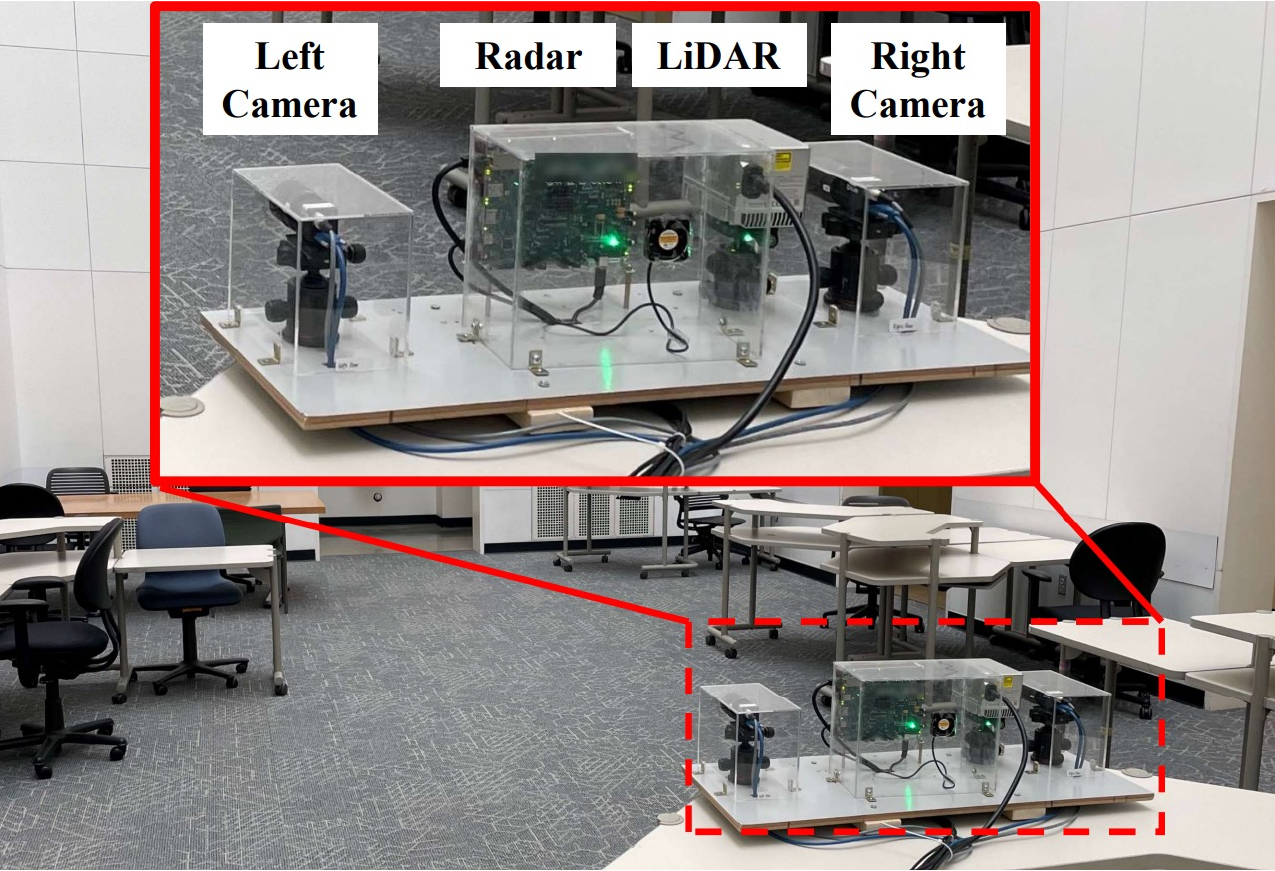}
\caption{Experimental hardware setup in an indoor environment for data collection.}
\label{fig:device}
\end{wrapfigure}
To construct the dataset, we recruit 10 participants to perform 6 types of actions. The experiments are divided into two categories. In the first category, participants are required to perform actions while standing, including waving, lifting legs, and random poses to increase the complexity of the actions. A video sequence of random poses involves a subject randomly performing one possible movement at a time, including stretching, bending, twisting, etc., each lasting 3 to 5 seconds. The second category focuses on walking, with additional actions added during the process, such as walking and waving or sitting down after walking. Each type of action is inherited from previous radar-based datasets \cite{xue2021mmmesh,sengupta2022mmpose,an2021mars}. Each sequence contains one action for 30 seconds, providing a variety of different difficulty levels. 
As a result, people can thoroughly evaluate their 3D HPE methods on a broad distribution of actions with diverse difficulties. 
Furthermore, we provide the dataset development kit to facilitate the usage and evaluation of future methods.
\begin{figure}[t]
\centering
\includegraphics[width=1\linewidth]{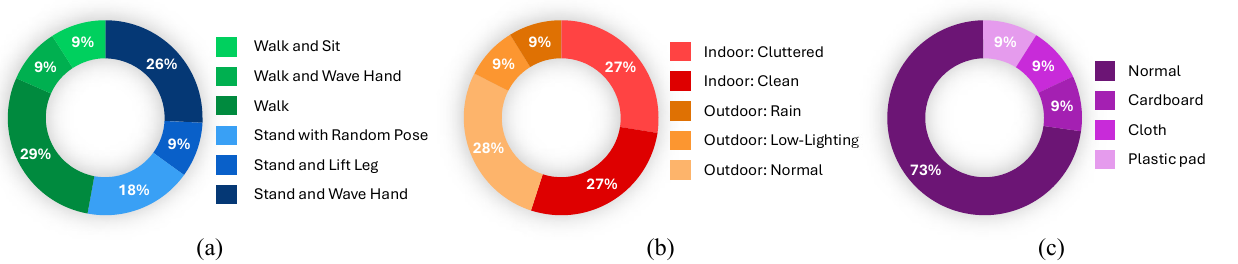}
\caption{Data distribution for RT-Pose dataset: (a) Activities; (b) Environmental conditions; (c) Occlusion conditions}
\label{fig:data_distribution}
\end{figure}

\subsection{Sensors}
The data collection hardware system comprises two RGB cameras, a non-repetitive horizontal scanning LiDAR, and a cascade imaging radar module, as shown in Figure~\ref{fig:device}. 
To extract human pose information from radar signals, the radar module's parameters have been specifically configured.
The radar operates at ten frames per second in this work. The radar module is equipped with 12 transmit (TX) and 16 receive (RX) antenna elements and operates in a multiple-input multiple-output (MIMO) mode~\cite{kim2021human}. 
This radar module provides the azimuth and elevation angle resolutions of 1.4 degrees and 18 degrees, respectively. Furthermore, this work implements a high-frequency slope with a long ramp time to efficiently utilize the hardware's sweep bandwidth, ranging from 77 GHz to almost 81 GHz. The longer sweep bandwidth of a chirp increases the radar range resolution~\cite{klauder1960theory,sengupta2020mm}, which is 4.5 cm in this work. Each frame consists of 64 chirps used for calculating radar Doppler signals, providing a velocity resolution of 3.9 cm/s. These carefully chosen configurations ensure the system's accuracy and reliability in capturing human poses.

\subsection{Data Collection}
\begin{figure}[t]
\centering
\includegraphics[width=0.9\linewidth]{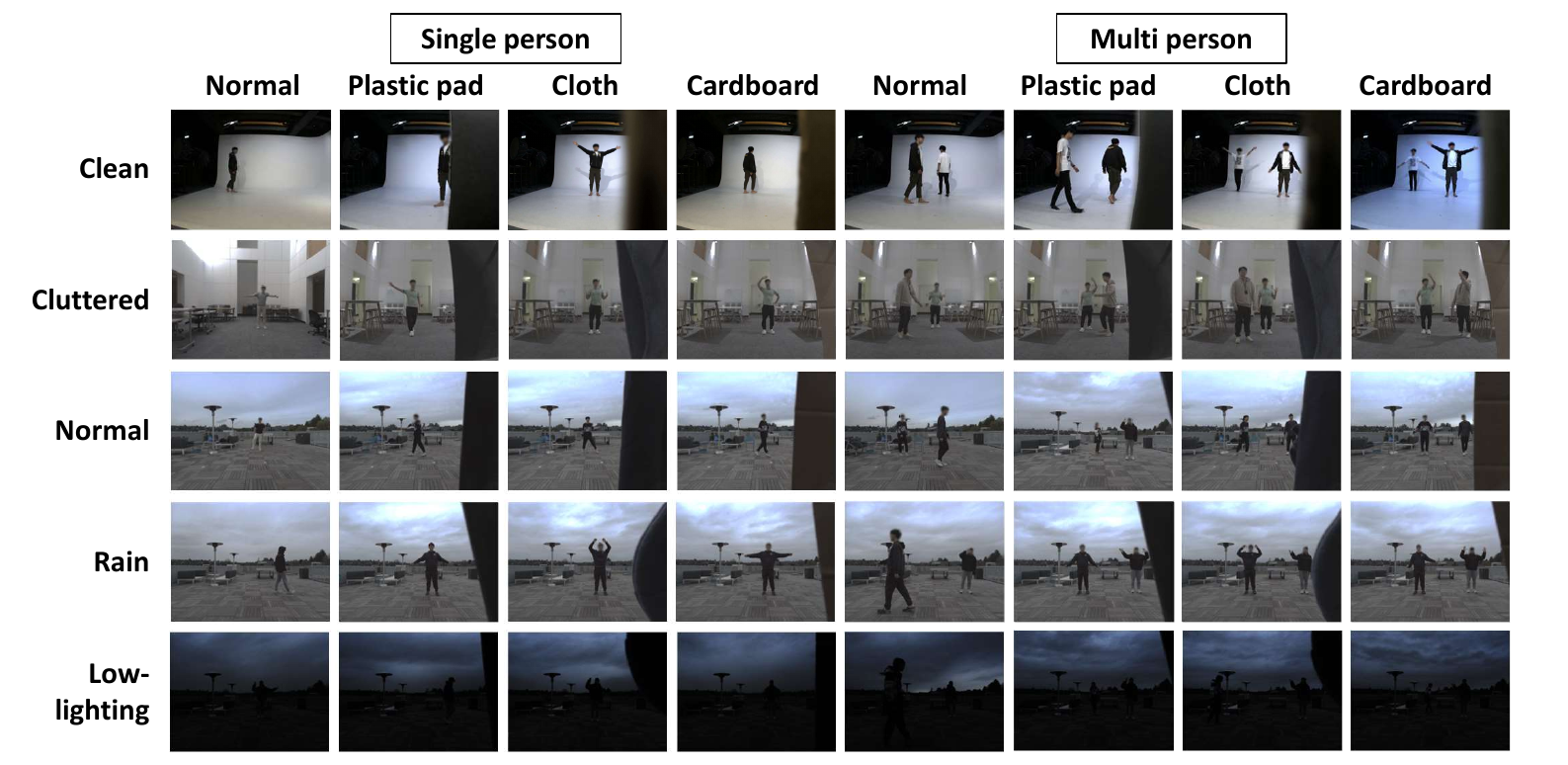}
\caption{Experimental instances across various indoor and outdoor conditions with diverse scenarios for data collection.}
\label{fig:env}
\end{figure}

We collect the dataset in 40 scenes with indoor and outdoor environments. 
Figure~\ref{fig:env} shows some of the collected data.
Indoor environments include clean and cluttered conditions, while outdoor ones include normal, rainy weather, and low-lighting conditions. Each experimental condition includes 8 scenarios, which are nonocclusive, and three occlusive conditions (cardboard, cloth, and plastic pad). 
Each nonocclusive and occlusive condition is recorded in single- and multi-person settings.
To consider the different fields of view (FoV) of all sensors, the sensor suite is situated 109 cm above ground, and an appropriate detection area is planned based on different scenarios. 
\subsection{Data Processing}
\begin{figure}[t]
    \centering
    \includegraphics[width=0.75\linewidth]{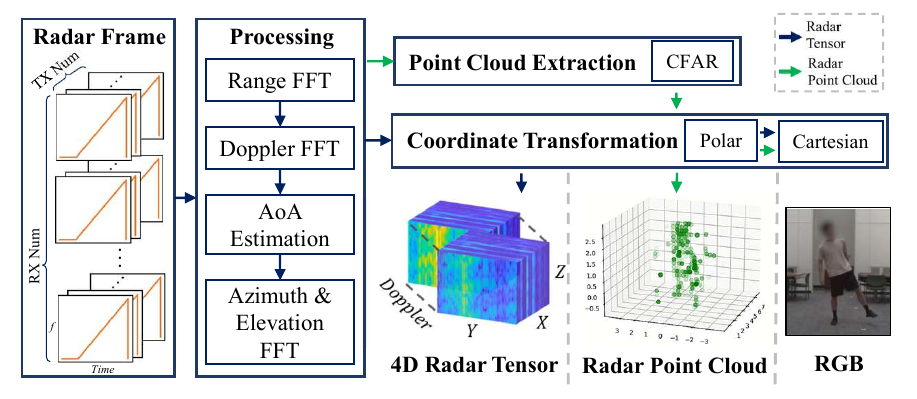}
    \caption{Radar signal processing flow. The green arrow line is for radar point cloud generation and the blue line is for 4D radar tensor generation. }
    \label{fig:radar_process}
\end{figure}
The captured radar signal, referred to as 4D tensor data, is processed by a series of steps \cite{heath2016overview, neemat2019reconfigurable, cruw3d}, as shown in Figure~\ref{fig:radar_process}. 
On the transmitter side, the frequency of the transmitted signals, or chirps, periodically increases based on the set frequency slope. 
Consequently, the received signal exhibits a different frequency than the transmitted signal. 
The frequency difference between transmitted and received signals can be used to estimate the range of the object reflecting the signal to the radar sensor by the Fast Fourier Transform (FFT). 
At the beginning of the data processing, each radar frame consists of 64 chirps, resulting in 64 range estimation results. Utilizing this range information, the velocity of the responding object can be measured through FFT, a phenomenon known as the Doppler effect. 
Second, the radar signal data is re-modulated according to the antenna position. This step is essential for the calculation of the angle of arrival (AoA), enabling the generation of highangular resolution results. 
Third, the radar signal from the azimuth and elevation directions also undergoes FFT processing to represent velocity responses in different directions. Finally, the entire radar data is transformed from polar coordinates to Cartesian coordinates. This transformation enhances the intuitiveness for the subsequent pose estimation. In this study, the processed 4D radar tensor has the dimension of $64\times32\times128\times256$, which correspond to velocity, the $z$-axis, the $y$-axis, and the $x$-axis, respectively.

\subsection{Annotation Workflow}

\begin{figure*}[t]
\centering
\includegraphics[width=0.75\linewidth]{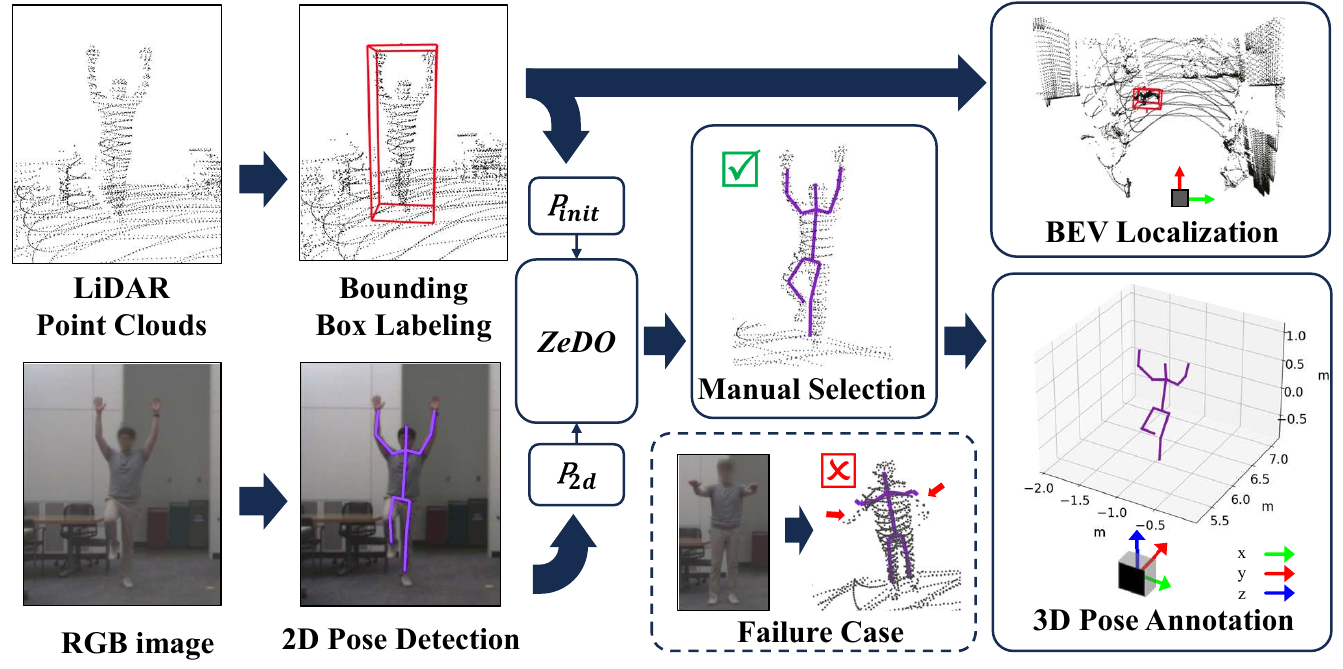}
\caption{Workflow of human localization and 3D pose ground truth annotations. Estimated 2D pose results, predicted by the pre-trained HRNet model, are denoted as \(P_{2d}\). The initial setting pose derived from LiDAR point clouds is denoted as \(P_{init}\). Both \(P_{2d}\) and \(P_{init}\) are inputs into ZeDO, an optimization-based pipeline for 3D pose estimation.}
\label{fig:annotation}
\end{figure*}
To achieve accurate human subject detection in a larger area with high-quality ground truth, our system integrates LiDAR and RGB camera data, as illustrated in Figure~\ref{fig:annotation}. The images provided by the camera are processed using the HRNet~\cite{sun2019deep} to extract 2D human poses. Then, a 3D HPE model, called ZeDO~\cite{jiang2024back}, utilizes the 2D poses to estimate the 3D poses as the initial 3D pseudo ground truth. However, estimating 3D poses using a monocular camera often lacks precision and stability in depth. Therefore, this work incorporates LiDAR for depth estimation of the human subject center. The LiDAR sensor captures environmental information and the subject's point cloud data, ensuring errors are under 2 cm inside the detection range of 20 m. The human subject is manually labeled with a 3D bounding box, which can provide an accurate 3D position.

ZeDO, a state-of-the-art (SOTA) method for cross-dataset HPE, implements an optimization-based pipeline for 3D pose estimation using 2D poses. This method requires 2D poses as input, and ZeDO is able to estimate multiple-hypothesis 3D poses. In this research, the center of the 3D bounding box provided by LiDAR is utilized as the pelvis of the initial pose, which enhances the accuracy of depth estimation compared to relying solely on the monocular camera data, and improves the performance of ZeDO. With the imported 2D pose from HRNet~\cite{hrnet}, ZeDO iteratively refines the 3D poses. To further improve annotation accuracy, an annotation tool is developed for manual selection and optimization of each frame to retain correct 3D human poses. The 3D HPE results from ZeDO will be put in the LiDAR point cloud for visualization, which is beneficial for humans to filter and modify the error pose annotation results. About 30\% of the data is removed or optimized after human correction, and this process ensures the quality of keypoints annotation. More details can be found in the supplementary material.

\section{HRRadarPose}
\begin{figure}[t]
\centering
\includegraphics[width=1.0\linewidth]{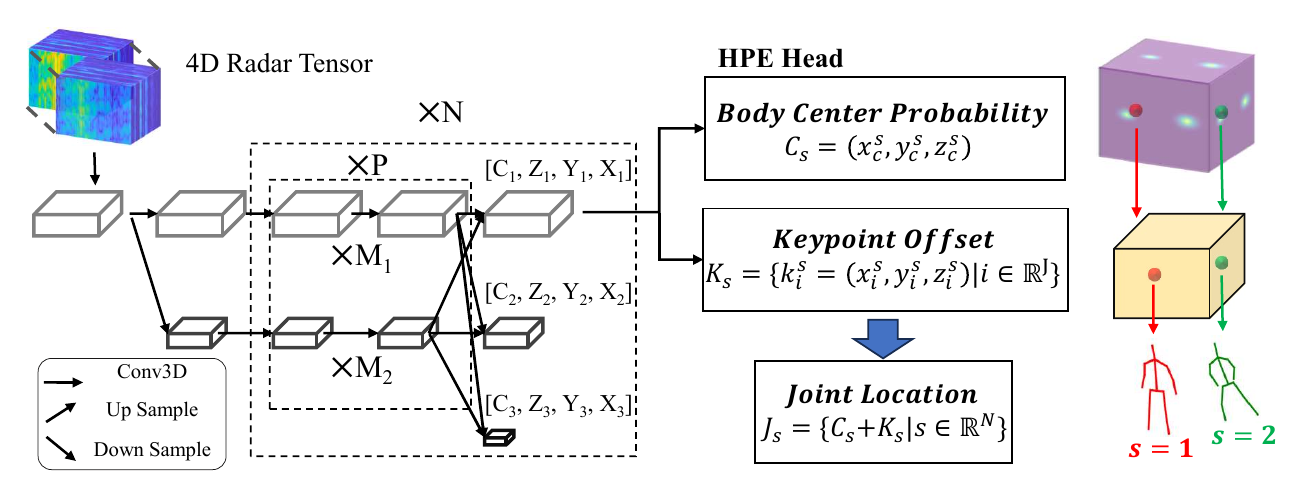}
\caption{
 HRRadarPose's architecture consists of a 3D convolutional backbone and a pose estimation head, which generates a body center's confidence map and joint keypoint offsets map. 
 To infer multi-person poses, we decode the prediction by adding the picked bodies' centers with their corresponding keypoint offsets. 
}
\label{fig:radarpose}
\end{figure}

Inspired by HRNet~\cite{hrnet}, renowned for its adeptness in extracting high-resolution representations, we build an architecture, HRRadarPose, with fully 3D convolutional layers.
By treating the Doppler axis, \(D\), as the input channels, our model extracts volumetric spatial features along the \(Z\), \(Y\), and \(X\) axes, representing vertical, horizontal, and depth directions in the Cartesian coordinate system. 
The HRRadarPose architecture is designed to preserve spatial resolution and assimilate semantic information from the 4D radar tensors, maintaining high-resolution representations and enabling the exchange of information between features of different resolutions during stage transitions.
Figure~\ref{fig:radarpose} illustrates an instance of the HRRadarPose structure with three stages. 
The structure starts with using 3D convolutional layers to transform the 4D radar tensors, \(T \in \mathbb{R}^{D \times Z \times Y \times X}\), into features, \(f_{in} \in \mathbb{R}^{C_{1} \times Z_{1} \times Y_{1} \times X_{1}}\). 
As the network proceeds from one stage to the next, it develops additional branches through strided convolution to widen the receptive field. 
Each branch \(i\) employs strided convolution to elevate its input feature \(f_{i}\) to a feature \(f_{i+1}\) in the succeeding branch \(i+1\), simultaneously doubling the channels and halving the spatial dimensions, \(f_{i+1} \in \mathbb{R}^{ 2C_{i} \times \frac{Z_{i}}{2} \times \frac{Y_{i}}{2} \times \frac{X_{i}}{2} }\).
Within each stage, a series of \(P\) modules execute \(M\) parallel convolutions. 
To align feature dimensions from different branches for information exchange, we use strided convolution followed by upsampling. 
At the end of the backbone, we take the feature with the highest resolution to serve as a unified representation for the pose estimation head. 
In the pose estimation head, two 3D convolutional branches are employed to process these features further. 
This bifurcation generates two outputs: a distribution map signifying the confidence of locating the human center and the location offset of joint keypoints. 
The right part of Figure~\ref{fig:radarpose} shows the decoding of the HRRadarPose network's output to multi-person poses.
The process starts with picking top-$k$ human centers, \(C_S\), with confidence scores over a threshold value. 
We denote these picked centers' indices as \(S\).
Then, we query the centers corresponding joint keypoints' offsets, \(K_S\), by looking up the output map of keypoint offsets head at \(S\). 
Lastly, we decode their joint keypoint locations, \(J_S\), by summing \(K_S\) and \(C_S\). 
In scenarios with multiple people, we apply range non-maximum suppression (NMS) to reduce overlapping detections. Our method is favorable in two folds:
The HRRadarPose's architecture gains efficiency through a single-stage workflow, avoiding complexity brought by region-proposal-based methods ~\cite{zhao2018rf}. In addition, our pose estimation head inherently learns per person's joint keypoints refraining from ambiguity of association between people's identity and joints. 

To train the center probability head, we use the pixel-wise focal loss as the classification loss, \( L_{class} \), defined as:
\begin{equation}
L_{class} =
\begin{cases}
    (1 - P_{xyz})^\alpha \log(P_{xyz}), & \text{if } Y_{xyz} = 1, \\
(1 - Y_{xyz})^\beta (P_{xyz})^\alpha \log(1 - P_{xyz}), & \text{otherwise},
\end{cases}
\end{equation} \ where \(Y_{xyz}\) is the three-dimensional human center Gaussian distribution map generated by pelvis locations, \(P_{xyz}\) is the probability of a person's presence at location, \((x,y,z)\), \(N\) is the number of foreground samples, \(\alpha\) and \(\beta\) are hyperparameters addressing class imbalance and focusing training on foreground samples.
To train the keypoint offset head, we construct the regression loss \(L_{reg}\) by taking the average of 15 keypoints' \(\mathcal{L}_{1}\), comparing the distance between predicted offsets and ground-truth offsets to the center location of a voxel in \(Y\) when \(Y_{xyz}\) equals to \(1\). 
The final loss is a weighted sum of \( L_{class} \) and \(L_{reg}\).

\begin{figure*}[t]
\centering
\includegraphics[width=0.9\linewidth]{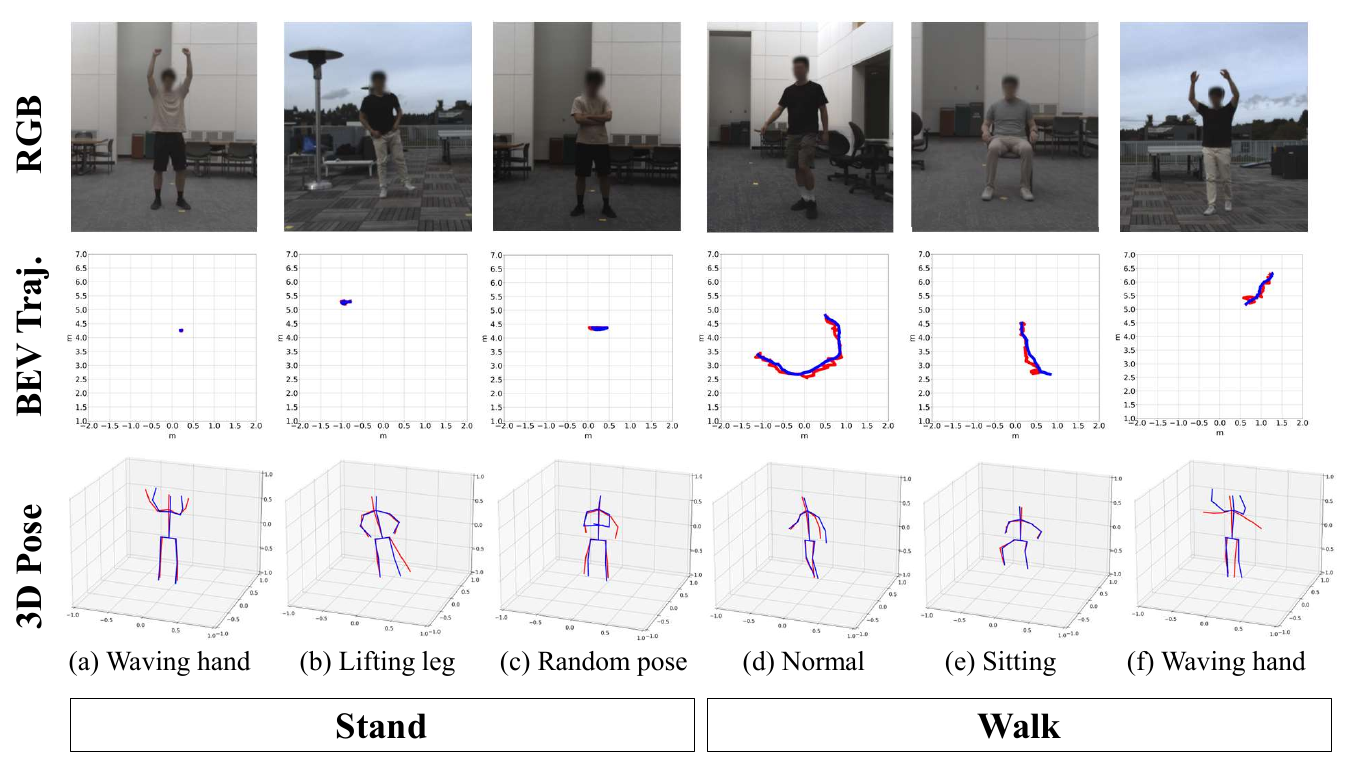}
\caption{Visualization of pose estimation results in test frames for six activities. The second row is the trajectory of the test sequence in BEV. The third row is the instance of the global 3D pose estimated result for each action. In both rows, the blue line is the ground truth, and the red line represents estimated results.}
\label{fig:action}
\end{figure*}
\section{Experiments} 
15 obvious keypoints of a human body are selected, aligning with the body skeleton model of the Human3.6M dataset~\cite{ionescu2013human3}. 
Each frame's 3D pose annotation in the sequence is manually validated to ensure the quality of the dataset. Figure~\ref{fig:action} demonstrates the testing results across various levels of action complexity, presenting the feasibility of 3D pose estimation using the proposed baseline.

\paragraph{Evaluation metrics}
We employ joint position error (JPE), which measures the mean of the Euclidean distance for every joint's predicted keypoints and the ground truth keypoints. Localization results are reported by the mean of the root position error (MRPE) ~\cite{wang2021deep}, which is the JPE of the pelvis keypoint. For the evaluation of 3D HPE, we use Mean per joint position error (MPJPE)~\cite{zheng2023deep} and absolute-MPJPE (Abs-MPJPE)~\cite{veges2019absolute}.

\paragraph{Baselines Comparison}
We evaluate the performance of the HRRadarPose in global localization and 3D HPE on the RT-Pose dataset. 
We compare our method with three baseline methods, as shown in Table \ref{tab:baseline}. mm-Pose \cite{sengupta2020mm} and mmMesh \cite{xue2021mmmesh}, utilizing radar point clouds for 3D HPE, are re-implemented to test on the RT-Pose dataset. 
Notably, mmMesh introduces a global localization model, combined with an RNN HPE module, showing more accurate estimation results in both MRPE and MPJPE compared to mm-Pose, which employs a simpler CNN model. RF-Pose 3D~\cite{zhao2018rf} uses the 4D radar tensor with a region proposal network to locate the subjects and estimate the 3D pose. 
The experimental results show that our HRRadarPose outperforms other baseline methods. On the RT-Pose dataset, we obtain  9.93 cm in MPJPE and 9.91 cm in MRPE. We find that radar-tensor-based methods reach less error than the methods using radar point cloud. 
We attribute this performance gap to complicated scenarios since it is difficult for point cloud-based methods to discriminate different key points in various poses. 
The results highlight the importance of using 4D radar tensors, which preserve raw and rich spatial-temporal information for the HPE tasks. 
An end-to-end architecture directly takes in 4D radar tensors to better face different human action scenarios in the real world.
\begin{table}[t]
\centering
\caption{Comparison of human localization and 3D pose estimation results of our HRRadarPose method and three baseline methods. $\dagger$ represents the radar point cloud-based methods.}
\resizebox{.7\linewidth}{!}{
\begin{tabular}{l|c | c c}
\toprule

 Method & MRPE (cm)             & MPJPE (cm)              & Abs-MPJPE (cm)          \\ \midrule
mm-Pose~$\dagger$~\cite{sengupta2020mm}          & 102.28                                & 21.26                   & 102.7               \\
mmMesh~$\dagger$~\cite{xue2021mmmesh}                                      & 62.69                                 & 13.78                   & 66.24                 \\
\midrule
RF-Pose3D~\cite{zhao2018rf}           &       29.17        &  18.43  &  38.05 \\
HRRadarPose~(Ours)     & \textbf{9.91}& \textbf{9.93}& \textbf{14.73}\\ 
\bottomrule
\end{tabular}
}
\label{tab:baseline}
\end{table}

\begin{table*}[t]
\centering
\caption{Comparison of the performance predicted by the HRRadar Pose in different complexity level activities, presenting different key points of the pose, the results of localization, and total 3D pose estimation results. 
The first group of rows lists the stand-related poses, while the other lists the walking-related poses. The complexity level increases from the top activity to the bottom within each group.}
\resizebox{1\textwidth}{!}{
\begin{tabular}{ll|cccc|c|cc}
\toprule
\multicolumn{2}{c|}{}    & \multicolumn{4}{c|}{JPE(cm)}  & \multicolumn{1}{c|}{} & \multicolumn{1}{c}{} & \multicolumn{1}{c}{} \\ 
\multicolumn{2}{c|}{\multirow{-2}{*}{Actions}} & \multicolumn{1}{c}{Thorax} & \multicolumn{1}{c}{Head} & \multicolumn{1}{c}{Ankle} & \multicolumn{1}{c|}{Wrist} & \multicolumn{1}{c|}{\multirow{-2}{*}{MRPE (cm)}}       & \multicolumn{1}{c}{\multirow{-2}{*}{MPJPE (cm)}}
& \multicolumn{1}{c}{\multirow{-2}{*}{Abs-MPJPE (cm)}}\\ \midrule
                             & Waving Hand     &             2.95               &               3.97&              3.61&              25.65&                 3.35&                 6.91&            7.97\\
                             & Lifting Leg     &              6.12&            7.92&             12.24&             8.88&             7.55&              7.06&              10.37\\
\multirow{-3}{*}{Stand:}      & Random Pose     &              7.78&           11.21&           11.75&                  32.12&                      7.67&                   12.47&                 15.12\\ \midrule
                             & Normal          &              4.31&         6.69&               13.52&               15.98&             13.67&            8.93&             16.92\\
                             & Siting     &          9.36&        13.03&                    13.93&             10.77&             14.11&                       9.44&                16.83\\
\multirow{-3}{*}{Walk:}       & Waving Hand          &             5.74&              10.37&             13.53&               43.16&                  12.82&    15.28&             21.22\\ \bottomrule
\end{tabular}
}
\label{tab:action}
\end{table*}

\paragraph{Action complexity Analysis}
RT-pose dataset provides six difficulty levels of actions to simulate real-world scenarios, enabling comprehensive evaluation of localization and 3D HPE results. 
We report the performance of the proposed HRRadarPose method in Table \ref{tab:action}. Generally, the joints along the human longitudinal axis, such as the thorax and the head, are more stable than those in the limbs. 
This phenomenon may be attributed to the larger reflective surface of the body's trunk compared to the limbs, providing a more reliable HPE based on radar signals. 

The MRPE of standing posture is below 7.67 cm, indicating that the pose complexity slightly influences the localization accuracy. Due to the chair's strong radar reflection, distinguishing between human subjects and the chair poses a challenge for localization. The MPJPE of walking and sitting is 9.44 cm, showing our HRRadarPose model can reliably estimate the sitting pose. However, the MRPE of this action sequence achieves 14.11 cm, demonstrating that sitting is a key factor for evaluating the performance of localization. The MPJPE of simple actions, such as standing and waving, standing and lifting legs, are around 7 cm, and the walking is 8.93 cm, demonstrating that our HRRadarPose model is comparable to other recently published radar pose estimate methods ~\cite{xie2023rpm,lee2023hupr}. However, the current model still encounters difficulty in estimating HPE in complex actions such as walking and waving. 
Therefore, the RT-pose dataset is a challenging benchmark, which can enable HPE models to realistically learn diverse pose scenarios.

\paragraph{Qualitative Result}

\begin{figure*}[t]
\centering
\includegraphics[width=0.8\linewidth]{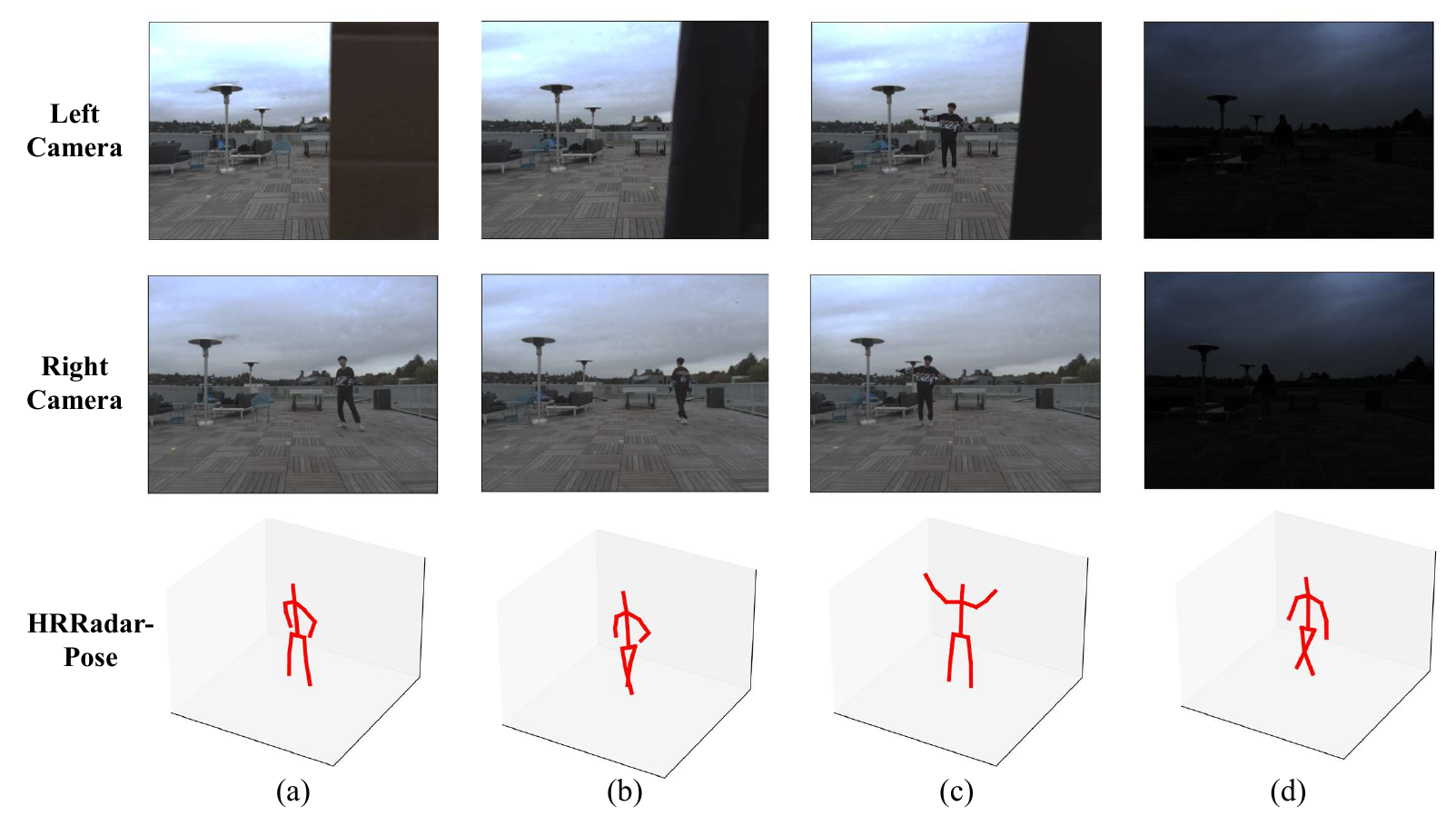}
\caption{HRRadarPose Results in different conditions. (a), (b), and (c) are occluded by cardboard, cloth, and plastic pad, respectively. (d) is in low-lighting conditions}
\label{fig:occ_vis}
\end{figure*}

Figure~\ref{fig:occ_vis} demonstrates the qualitative results of different occlusion conditions and a low-lighting instance.
Due to the hardware setting of our system, supporting synchronized two camera data, the subject's pose can be visualized by another camera if one of them is occluded.
Therefore, the 3D annotation results can be generated even on occlusion conditions. 
Our method provides reliable pose estimation results even when the radar module is blocked or with low-lighting conditions.
Table~\ref{tab:Occ} shows the HPE results of occlusion in normal outdoor conditions. Although the radar module is occluded, the mmWave signal is still robust enough as input for the HRRadarPose model to estimate HPE, demonstrating the characteristics of radar and the importance of using radar for HPE.

\paragraph{Ablation Study}

To illustrate the advantage of modeling motion features using Doppler information by our HRRadarPose, we remove Doppler information by averaging 4D radar tensors along the Doppler axis and train the models with and without Doppler information using the same setup. 
Table~\ref{tab:input} shows that with the inclusion of Doppler information, our HRRadarPose enjoys a performance gain due to its capability to extract motion features, enhancing keypoint localization.
\begin{table}[t]
  \centering
  \begin{minipage}[c]{0.48\textwidth}
        \centering
        \caption{ HPE results from HRRadarPose in normal and occlusive scenarios.}
        \resizebox{0.9\linewidth}{!}{
        \begin{tabular}{c| c c}
        \toprule
        
         Occlusion & MPJPE (cm)              & Abs-MPJPE (cm)          \\ \midrule
               & 9.84                  & 14.52               \\
        \checkmark                     & 11.57                   & 18.55                 \\
        \bottomrule
        \end{tabular}
        }
        \label{tab:Occ}

  \end{minipage}
  \hspace{0.02\textwidth}
  \begin{minipage}[c]{0.48\textwidth}

    \centering

    \caption{Ablation Study of Using Dopper information.}
    \resizebox{0.9\linewidth}{!}{
    \begin{tabular}{c|cc}
    \toprule
    Doppler           & MPJPE~(cm) & Abs-MPJPE~(cm) \\ \midrule
                          &                   11.31                         &                         22.34                       \\
    \checkmark                             &               9.91                             &                  14.73\\ 
    
    \bottomrule
    \end{tabular}
    }
    \label{tab:input}

  \end{minipage}
\end{table}


\begin{table}[t]
\centering
\caption{Our Radar-Doppler Convolutional Block in HR-Net.}
\resizebox{0.65\linewidth}{!}{
\begin{tabular}{c|cc}
\toprule
HRNet-Block-Design           & MPJPE~(cm) & Abs-MPJPE~(cm) \\ \midrule
        Traditional Block (RGB-Vision)              &        10.35                                     &                17.59                                \\
        Ours (Radar-Doppler)                   &         9.93                                   &          14.73        \\ 

\bottomrule
\end{tabular}
}
\label{tab:hrnet_comp}
\end{table}

\begin{table}[t]
    \centering
    \begin{minipage}[c]{0.48\textwidth}
        \centering
        \centering
        \includegraphics[width = 0.7\linewidth]{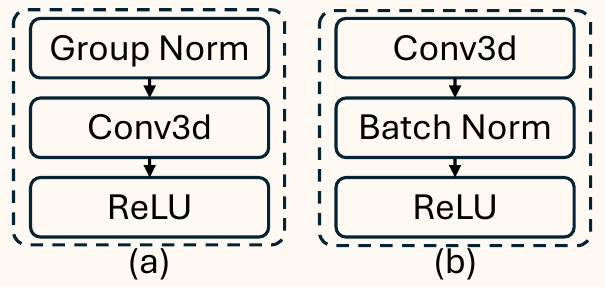}
        \captionof{figure}{Illustration of two variants of CNN blocks in HRNet. (a) Ours (Radar-Doppler). (b) Traditional Block (RGB-Vision).}
        \label{fig:block}

    \end{minipage}
    \hspace{0.02\textwidth}
    \begin{minipage}[c]{0.48\textwidth}
        \centering
        \caption{
        Comparison of common 4D radar tensor-based HPE head and ours.
        }
        \resizebox{0.9\linewidth}{!}{
        \begin{tabular}{l|cc}
        \toprule
        HPE Head           & MPJPE~(cm) & Abs-MPJPE~(cm) \\ \midrule
                   RF-Pose3D~\cite{zhao2018rf}                  &                              15.52              &               18.08                                 \\ 
        Ours                       &             10.24                               &                     17.63                          \\ 
        \bottomrule
        \end{tabular}
        }
        \label{tab:head}
    \end{minipage}
\end{table}
Compared with RPM 2.0, our input data and backbone design are different. 
The proposed HRRadarPose network extracts better feature representation from 4D radar tensors through modified convolutional blocks. 
As shown in Figure~\ref{fig:block}, we replace the batch norm with the group norm of convolutional blocks in HRNet. Since we treat the channel-wise features as a Doppler axis, the group norm inherently handles the feature in different speed groups, which normalizes the data effectively. 
Table~\ref{tab:hrnet_comp} shows the results, proving that the modified CNN block is more suitable for Doppler radar than the original HRNet.
Compared with our method, most previous work did not leverage Doppler information ~\cite{xie2023rpm,yu2023mobirfpose}. 
We validate that motion features ~\cite{huang2023observation,lin2015position} are crucial for aiding deep learning models in capturing more human body characteristics over the environment. 

To ablate our design choice of HPE heads, we use the same backbone, replace our pose estimation head with RF-Pose3D's~\cite{zhao2018rf} head, and train both models for the same epochs. 
RF-Pose3D formulates the HPE as a multi-keypoints classification problem, where their model outputs confidence distributions of every human body joint. 
As shown in Table~\ref{tab:head}, our design is favorable in joint keypoints localization in the metric space, indicated by a lower Abs-MPJPE. 
Moreover, lower MPJPE from our HPE head's result validates stronger inter-joint relationships modeling capability.

\section{Limitation and Future Work}
First, to leverage the input data types for training the 3D HPE model, the computational resource is a crucial part. 4D radar tensor in Cartesian coordinate consumes up to 100 MB per frame, which  decreases the training speed. 
Second, we limit the experimental data collection scope from 2 m to 8 m because the camera is hard to clearly capture 2D pose skeletons at a large distance and LiDAR can not easily collect completed point clouds in near vision due to the constraints of FoV.
Third, the proposed baseline HRRadarPose lacks robustness in accurately estimating poses during complex activities. Capturing human pose accurately on 4D radar tensor remains an unresolved challenge.

\section{Conclusion}
In this paper, we propose RT-Pose, the first dataset for human pose estimation (HPE) with synchronized and calibrated 4D radar tensors, LiDAR point clouds, and RGB images. 
RT-Pose provides cluttered scenarios and human activities of different complexity levels to enhance the difficulty of pose estimation for practical applications. In addition, having various modalities is beneficial for our semi-automatic 3D pose annotation optimization process. The proposed HRRadarPose is the first single-stage architecture designed for estimating human pose using 4D radar tensors as the input. The results indicate that the proposed HRradarPose outperforms previous works, in which datasets are collected with simpler actions and cleaner scenarios. Our work contributes to offering a three-modality dataset with 4D radar tensors for HPE, which is promising to be applied for complex actions and different scenes. 
We hope this work encourages future development of 4D radar-based HPE methods. 
\clearpage

\section*{Acknowledgements}
We thank Yizhou Wang \footnote{http://yizhouwang.net/} for helping us build the data collection system and the radar data processing pipeline.
We thank Hou-I Liu \footnote{k39967.c@nycu.edu.tw}, Cheng-Yi Huang \footnote{n28121521@gs.ncku.edu.tw}, Yi Shiang Chen\footnote{m16121051@gs.ncku.edu.tw},  Chi-Yuan Chan\footnote{n26120587@gs.ncku.edu.tw}, and Yu-Hsien Lu\footnote{n26120587@gs.ncku.edu.tw} for helping us collect the RT-Pose dataset. 
We thank the National Science and Technology Counci of the Republic of China, Taiwan, under Projects of NSTC 112-2917-I-006-007 for its funding support.

\bibliographystyle{splncs04}
\bibliography{main}
\end{document}